\documentclass[letterpaper, 10 pt, conference]{ieeetran} % Comment this line out if you need a4paper
\pdfoutput=1

\IEEEoverridecommandlockouts % This command is only needed if you want to use the \thanks command
\usepackage[utf8]{inputenc}
\usepackage{graphicx}
\usepackage{float}
\usepackage{amsmath}
\usepackage{array}
\usepackage{setspace}
\usepackage{pgf}
\usepackage{eso-pic}
\usepackage[hidelinks,bookmarks=false]{hyperref}

\usepackage{tikz}
\usepackage{textcomp}

\newcommand\copyrighttext{\footnotesize \textcopyright 2016 IEEE. Personal use of this material is permitted. Permission from IEEE must be obtained for all other uses, in any current or future media, including reprinting/republishing this material for advertising or promotional purposes, creating new collective works, for resale or redistribution to servers or lists, or reuse of any copyrighted component of this work in other works.}
\newcommand\copyrightnotice{%
\begin{tikzpicture}[remember picture,overlay]%
\node[anchor=south,yshift=10pt] at (current page.south) {\fbox{\parbox{\dimexpr\textwidth-\fboxsep-\fboxrule\relax}{\copyrighttext}}};%
\end{tikzpicture}%
}

\allowdisplaybreaks

\newlength{\OLDtextheight}
\title{\LARGE \bf
Real-Time Image Distortion Correction: Analysis and Evaluation of FPGA-Compatible Algorithms
}

\author{Paolo Di Febbo$^{1}$, Stefano Mattoccia$^{2}$, Carlo Dal Mutto$^{3}$% <-this % stops a space
\thanks{$^{1}$DISI, University of Bologna, Italy 

	{\tt\small paolo.difebbo1 at gmail.com}}%
\thanks{$^{2}$DISI, University of Bologna, Italy

	{\tt\small stefano.mattoccia at unibo.it}}%
\thanks{$^{3}$Aquifi, Inc.
	{\tt\small cdm at aquifi.com}}%
}

\begin{document}

\maketitle
\thispagestyle{empty}
\pagestyle{empty}

%%%%%%%%%%%%%%%%%%%%%%%%%%%%%%%%%%%%

\begin{abstract}

Image distortion correction is a critical pre-processing step for a variety of computer vision and image processing algorithms. Standard real-time software implementations are generally not suited for direct hardware porting, so appropriated versions need to be designed in order to obtain implementations deployable on FPGAs. In this paper, hardware-compatible techniques for image distortion correction are introduced and analyzed in details. The considered solutions are compared in terms of output quality by using a geometrical-error-based approach, with particular emphasis on robustness with respect to increasing lens distortion. The required amount of hardware resources is also estimated for each considered approach.

\end{abstract}

%%%%%%%%%%%%%%%%%%%%%%%%%%%%%%%%%%%%

\section{Introduction}
\copyrightnotice
Image distortion correction is a fundamental task in several practical computer vision and image processing applications. Among them, stereo vision is a very common scenario. Many single-camera applications also require distortion correction. Examples of such applications are: ego-motion estimation \cite{a_robust_method_ego_motion}, metric measurements \cite{metric_measurements}, food-portion size estimation \cite{food_portion_size_estimation}, image stitching \cite{mosaicing}, and many more.

A large number of these applications are real-time and mobile-oriented, requiring therefore low latency and low-power image processing. Implementing distortion correction algorithms on specialized or reconfigurable devices can be an optimal solution to reduce latency and power consumption, as well as to unload the CPU from this task. In particular, Field Programmable Gate Arrays (FPGAs) are often the preferred choice when aiming at maximizing the performance/power-consumption ratio. Moreover, analyzing a typical vision processing pipeline, image distortion correction is often performed immediately after image acquisition from raw sensor(s). Since most common embedded systems target the entire vision pipeline or its portion on specialized or reconfigurable devices, it is implied that image distortion correction is almost mandatory to be implemented on such hardware rather than on a GPU in order to avoid ineffective transfers back and forth from the shared memory.

Nevertheless, porting image distortion correction algorithms to specialized hardware is not straightforward. Even if locality is assumed in the distortion correction process, there are still parts of the algorithm which cannot be ported naively from their software implementation counterpart. For instance, the distortion correction function which relates input and output images is computationally complex. While a Look-up-Table (LUT) approach is the ideal solution in software implementations, it might be too demanding in terms of memory when it comes to FPGA implementation. Different solutions have been proposed in the literature in order to overcome this problem, but they are rarely compared among each others. Therefore, in practical implementations, the decision of choosing one approach instead of the other is often made without considering the full spectrum of options. This may lead to over-sized, over-consuming, inefficient, expensive systems, which often provide results comparable to the ones enabled by more efficient but less popular alternatives. The scope of this paper is to clarify the current status of image distortion correction on FPGA, by reviewing and evaluating different solutions in terms of image quality and hardware resources. In particular, we evaluate the robustness of state-of-the-art approaches with respect to increasing lens distortions as well as to their hardware footprint. Both these factors have to be considered in order to determine the best trade-off between resources usage and output quality.

In the first part of the paper, an overview of the image distortion correction algorithm is given and its standard software implementation is explained. In the second part, an appropriate hardware (FPGA) design is described, with particular focus on the real-time output capability and memory usage optimization. The main hardware modules used within the design are described with a particular focus on the most critical component, \textit{i.e.}, the pixel-to-pixel \textit{remapping function} module. Concerning this task, we consider the de-facto standard distortion model proposed by \cite{heikkila, a_flexible_new_technique} and implemented in \cite{bouguet} and in the OpenCV library \cite{learning_opencv,initundistortrectifymap}, which accounts for both \emph{radial} and \emph{tangential} distortion. Similar reasoning can also be applied to more complex models, such as the \emph{rational} one proposed by \cite{a_rational_function_lens}, which however is less commonly adopted.

Possible approximated versions of the standard software-implemented approach are defined in order to meet the hardware constraints, and compared by analyzing their geometric error according to the reference software map. Along with hardware resources utilization estimates for each different approach, this study allows to determine which of these solutions could be the best choice in relation to different lenses and target FPGA architectures.

It is important to state that the proposed evaluation is built over the specific distortion model chosen \cite{initundistortrectifymap}, and conclusions might be different if simpler models are considered. This model has been selected since it is very commonly used and general purpose, covering most of the typical scenarios for both stereo vision and single camera image processing. % In conclusion, the actual resources usage of our specific implementation will be described.

\section{Related Work}

Distortion correction algorithms have been subject of several studies in the literature. In regards to the specific mathematical models for defining the distortion geometry, different models have been introduced \cite{heikkila,a_flexible_new_technique,bouguet,a_rational_function_lens}. Determining the distortion of the particular lenses requires a calibration stage, in which the image information from the camera is used to estimate the parameters for the distortion model through different techniques \cite{a_flexible_new_technique,straight_lines_have_to}. The distortion correction methodologies considered in this paper are agnostic to the calibration approach deployed.

Image distortion correction has been implemented in a wide variety of different contexts. The well-known OpenCV library \cite{learning_opencv} includes probably one of the most popular software versions of the distortion correction algorithm \cite{initundistortrectifymap}. It has also been adopted within GPUs \cite{computer_vision_on_the_gpu} and FPGAs  \cite{a_passive_rgbd_sensor_for_accurate}. Focusing on FPGA implementations, many alternatives have been designed for overcoming the problems mentioned earlier about the input/output mapping function complexity. Several papers adopt a straightforward porting of the software Look-up-Table version \cite{lut_based_image_rectification,a_real_time_embedded_system}. Given the high amount of memory needed in these approaches, external dynamic memories are required, leading to bulky and expensive hardware systems when such memory is not required for other purposes. Other approaches compute the mapping function \emph{on-the-fly} by implementing the complete camera and distortion model functions \cite{initundistortrectifymap} on hardware \cite{fpga_design_and_implementation,fpga_based_real_time_visual_servoing}. Although this approach is very accurate, it may be expensive in terms of gates and also redundant or inefficient from the point of view of power consumption, since the same model computation is performed repeatedly, from scratch, for each point of each frame. Another solution is based on a \emph{sub-sampled} version of the full resolution LUT method, using simple bilinear interpolation to get the specific pixel-to-pixel mapping value. This solution has been considered less frequently \cite{real_time_rectification_for_stereo_correspondence,high_resolution_stereo_in_real_time} than other approaches, even though it provides an effective trade-off between resources and performance.

\section{Image Distortion Correction Algorithm}

The distortion correction algorithm takes a raw camera image as input, and produces as output a \emph{distortion-corrected} version of the same image, according to the camera parameters estimated at calibration time. The image distortion correction algorithm warps the input image such that the lens distortion effects are compensated and it optionally performs a virtual rotation of the camera frame according to the specific usage needs. This latter transformation is needed for example in case of image rectification for stereo vision, where the images from the camera pair are required to appear aligned as if they were actually acquired by a perfectly aligned system. This setup is often referred to as \emph{standard form}.

Usually, a pixel-to-pixel mapping function is used to relate every pixel of the output (distortion-corrected) image to a pixel (located with sub-pixel accuracy) of the input distorted image, according to the specific correction to apply that is determined at camera calibration time. This design allows to easily compute each output pixel value by simply retrieving its corresponding pixel from the input image through the mapping function; this specific part of the algorithm is called \emph{image remapping}.

%\begin{figure}[t]
%	\centering
%	\includegraphics[width=6.0cm]{images/undistort.jpg}
%	\caption{Example of distorted input image (left) and distortion-corrected output image (right).}
%	\label{rectified}
%\end{figure}

\subsection{Image Remapping}

Image distortion correction can be regarded as a particular version of a more generic algorithm, that is image remapping. The main idea behind image remapping consists in defining a coordinates relationship between the input (\texttt{src}) and the output (\texttt{dst}) pixels through two mapping functions (\textit{i.e.}, $map_x(\cdot,\cdot)$, $map_y(\cdot,\cdot)$):
\begin{align}
\texttt{dst} (x,y) =  \texttt{src} (map_x(x,y),map_y(x,y)) \label{dst_src_mapping}
\end{align}
in which $map_x(x, y)$ is the mapping function relative to the $x$-component and $map_y(x, y)$ is the mapping function relative to the $y$-component.

Considering Equation (\ref{dst_src_mapping}), and given a pre-defined mapping function ($map(\cdot,\cdot)$), the output image \texttt{dst} can be computed according to the following two steps. First, for each output pixel $(x, y)$, the value of $map_x(x,y)$ and $map_y(x,y)$ is retrieved according to an appropriate approach, with $map_x(x,y)$ and $map_y(x,y)$ generally encoded with floating point values. Then, the algorithm gets the corresponding pixel value from the input image \texttt{src}, and stores it in the output pixel \texttt{dst}$(x, y)$. Since the source image coordinates $map_x(x,y)$ and $map_y(x,y)$ are not necessarily integers, the desired source pixel is actually somewhere in between four source pixels, therefore it is required to perform a bilinear interpolation of these pixel values according to the map function.

The important part of the image distortion correction algorithm, which characterizes it compared to a generic image remapping algorithm, is the specific mapping function $map(\cdot,\cdot)$. The $map_x(\cdot,\cdot)$ and $map_y(\cdot,\cdot)$ functions are computed according to the considered distortion model \cite{initundistortrectifymap}, such that the input image -- after remapping -- will actually appear distortion-corrected and possibly rotated.

\subsection{Standard Software Implementation}

In order to understand how the distortion correction algorithm works, and what are the main problems when it comes to its implementation on FPGA, it is convenient first of all to analyze its software implementation. In this paper, the OpenCV implementation is considered as a reference \cite{learning_opencv, initundistortrectifymap}.

The mapping function is usually defined according to a \emph{pinhole} camera model which includes radial and tangential lens distortion, along with a rotation operation for possible use when required. In this paper, we consider the commonly used model of \cite{heikkila, a_flexible_new_technique}, implemented in \cite{bouguet, learning_opencv}. The actual function is fully described in \cite{initundistortrectifymap}.

In the OpenCV implementation, this mapping function is computed beforehand -- for each pixel $(x, y)$ -- and all of its values are stored in memory. The remapping method accesses the pre-computed map values at run time in a Look-up-Table fashion. Although this is a very efficient solution in software, since the map function is computed only once at initialization time, this might be not feasible on a standard sized FPGA without external memory due to the size of the full-resolution LUTs.

\section{Image Distortion Correction on FPGA}

This section provides a description of our distortion correction algorithm for FPGA. This design is aimed at maximizing throughput and clock frequency through a deep datapath pipelining, and minimizing resources usage, both in terms of memory footprint and gates usage, through appropriate design techniques. Both goals have been successfully achieved, allowing real-time distortion correction at a pixel clock frequency of 100+ MHz. This easily enables processing of 720p video streams at 60+ Hz. % VGA resolution at 60 fps is currently used.

In this section, a detailed hardware description of the inverse mapping function for distortion correction (\emph{Map Module} in Figure \ref{Design-Overview}) is omitted, since its adaptation from the standard software implementation is not straightforward and a further explanation is required. Therefore, Section \ref{FPGA_REMAPPING} is focused on this specific problem analyzing in detail the constraints enforced by each examined solution to the underlining hardware design. 

\subsection{Top Level Architecture and Sub-modules}

\begin{figure}[t]
	\centering
	\includegraphics[width=7.0cm]{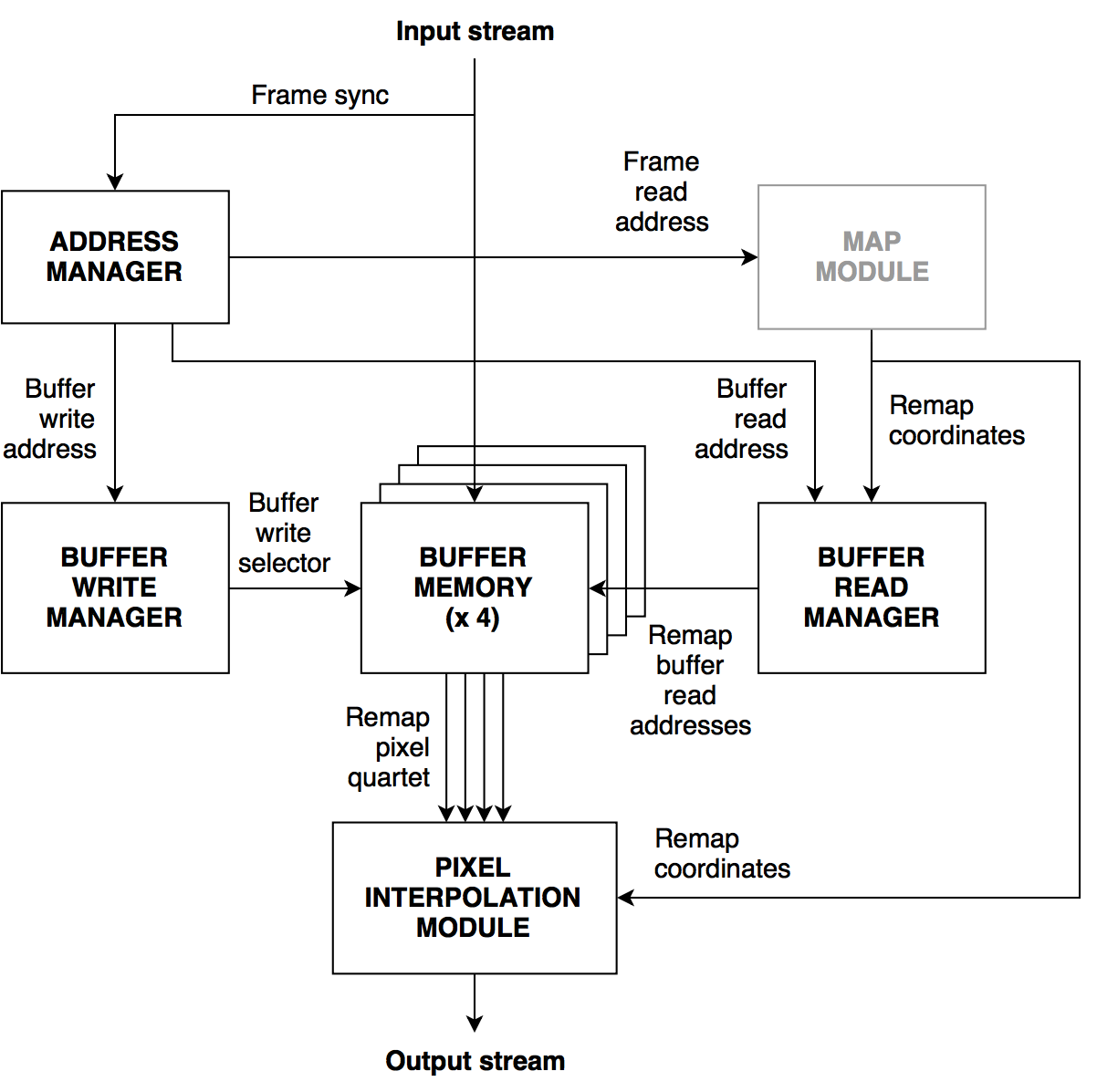}
	\caption{Overall design of the image distortion correction algorithm on FPGA}
	\label{Design-Overview}
\end{figure}

The high level block diagram of this design is outlined in Figure \ref{Design-Overview}. Incoming image data is written into a circular buffer, whose size is characterized by a pre-defined number of image rows. This amount of rows should be depending on the minimum and maximum \emph{relative vertical displacement} values of the specific distortion correction map that is currently being used. On the \emph{read} side of the buffer, with a delay -- in number of lines, related to the input frame start -- that corresponds at least to the maximum \emph{relative vertical displacement} value, the output image generation begins using the same timing specs as the input one. In this way, the buffer always contains all the pixels of the input image that are required for remapping purposes.
For the most important modules in Figure \ref{Design-Overview} we provide below a brief description of their functionality.

\begin{figure}[t]
	\centering
	\includegraphics[width=3.5cm]{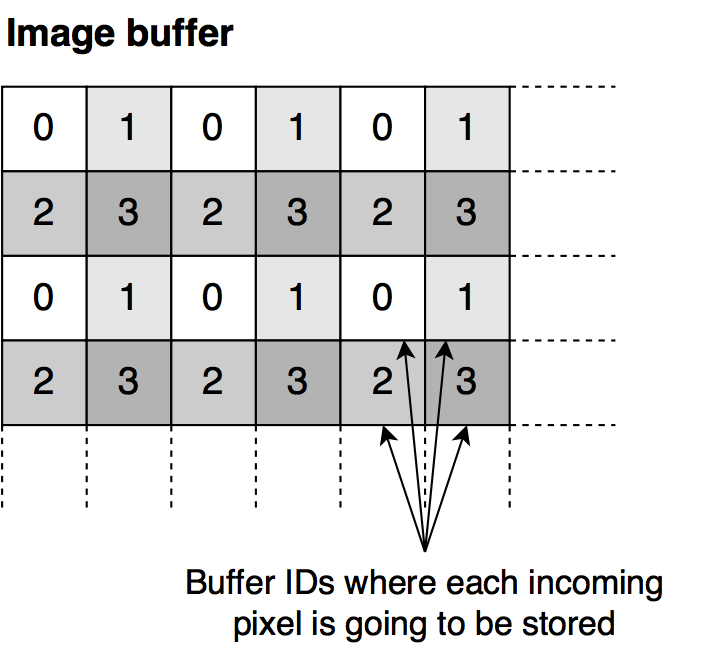}
	\caption{Interleaved data structure enabling parallel access to four pixels. Each number corresponds to the specific \emph{Buffer Memory} (Figure \ref{Design-Overview}) in which that pixel is stored}
	\label{buffer_pattern}
\end{figure}

\subsubsection{Buffer Write Manager}

This module manages the writing mechanism that allows every input pixel to be written into one of the four different buffer memories, according to a strategy that makes possible to access anytime (from the reading point of view), in the same clock cycle, four different pixels that are needed for an interpolation as depicted in Figure \ref{buffer_pattern}. For any interpolation quartet within the input image, its four pixels can be accessed in parallel, since they are always stored in different buffers.

\subsubsection{Buffer Read Manager}

This module combines the current \emph{Buffer read address}, generated by the address manager, with the remapping function coordinates that are related to that current output pixel, in order to compute the buffer addresses containing the input pixels that are needed to fill the current output pixel. Specifically, the remap coordinates are provided by the map module as \emph{relative} coordinates, meaning that they represent the relative position of the input distorted pixel with respect to the current output pixel:
\begin{align}
\nonumber \texttt{dst} (x,y) =  \texttt{src} (x + map_x(x,y),y + map_y(x,y))
\end{align}

\subsubsection{Pixel Interpolation Module}
This module, given the four neighbor pixels that are coming from the buffers in parallel at the same clock cycle, computes their bilinear interpolation according to the sub-pixel map value.

\section{Mapping Function on FPGA}
\label{FPGA_REMAPPING}

\begin{figure}[t]
	\centering
	\includegraphics[width=8.8cm]{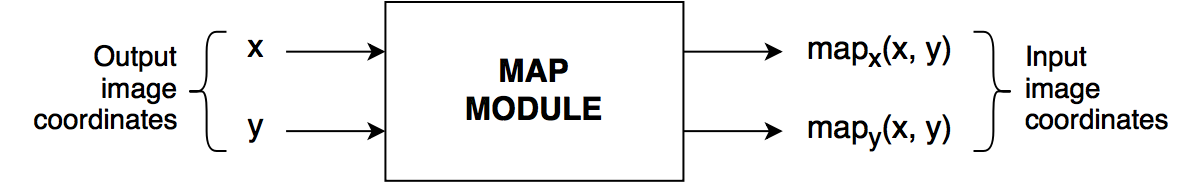}
	\caption{Detail of the remapping function on hardware}
	\label{map-module}
\end{figure}

Within our FPGA design discussed in the previous section, the task of determining the input (distorted) coordinates from the output image coordinates is assigned to the \emph{Map Module} depicted in figures \ref{Design-Overview} and \ref{map-module}. Possible issues arise when implementing the distortion correction mapping function on FPGA. The most challenging problem is that a conventional software implementation of the image distortion correction algorithm uses pre-computed full resolution maps that are entirely stored in memory as Look-up-Tables. Even considering low resolution VGA frames, the complete map size is too big for being stored in static memories (block ram/rom). Therefore, if external memory devices (e.g., DDR, SRAM) are not included in the hardware design, the methodology borrowed from the conventional software approach is not feasible for hardware implementation. Moreover, this solution requires a significant amount of memory bandwidth.

It is then required to find alternative solutions to the conventional Look-up-Table approach adopted for typical software implementations. Different techniques can be used in order to achieve different goals, that include primarily: robustness to the lens distortion with accuracy comparable to the conventional approach, reduced FPGA resource utilization and small memory footprint compatible with amounts available even in cheaper FPGAs.

\subsection{Complete On-the-fly Function Computation}

A first hardware-compatible approach that comes into mind by looking at the considered remapping function \cite{initundistortrectifymap} is the one of computing from scratch each $(map_x(x,y), map_y(x,y))$ value from the current $(x, y)$ coordinates. This approach, which avoids to store the full pre-computed map, is usually called \emph{on-the-fly} map computation \cite{fpga_design_and_implementation,fpga_based_real_time_visual_servoing}, completely avoids the need of memory, but has some drawbacks.

First of all, for each input frame within the same stream session, these values are computed each time for scratch, even if they always have the same exact value for the same coordinates among frames. This introduces redundant computational costs, which are directly depending on the complexity of the distortion model. Moreover, the floating point computation required by this approach might not be available or too demanding in terms of resource utilization and thus an approximated fixed point computation is mandatory in most cases. This strategy might result in an approximation error that is not negligible in comparison with the resolution of the standard software approach. Therefore a further analysis, provided in the experimental results section, is needed in order to better understand how accurate the fixed point computation should be, in terms of fractional bits, in order to preserve the integrity of the mapping. Despite these facts, most of the image distortion correction approaches on FPGA adopt this strategy.

\subsection{Map Sampling}

As outlined before, the main problem that the standard software solution faces when it needs to be designed on hardware is the size of the full resolution Look-up-Table for the remapping function. Different implementations within the current literature ``force'' to port this exact solution on hardware by using a complete software-like Look-up-Table \cite{lut_based_image_rectification,a_real_time_embedded_system}. However, because of the size of such structure, an external memory is needed in order to store it, making the design more complex, expensive, and demanding in terms of memory bandwidth.

An alternative approach, that comes to mind after this preface, is the one of using a size-reduced version of the Look-up-Table. The software computed pixel-to-pixel LUT can be subsampled, generating a lower resolution version of it, by picking one value every $2^n$ values for both axes. Bilinear interpolation of these samples can be used to reconstruct an approximation of the specific pixel's map value. The approximation error gets bigger when $n$ increases, \emph{i.e.}, when using less samples.

However, as reported in the next section, the subsampling factor $n$ can be significantly increased -- making the actual sampled map very small -- without compromising too much the quality of the map values. This approach is very convenient in terms of hardware resources since it requires only two simple bilinear interpolators (one for the $x$ map coordinates, one for the $y$ ones), plus a small amount of internal memory to store the subsampled map.

Moreover, even if this approach could be often considered as the best trade-off for the hardware map function problem, it is much less frequently adopted compared to the on-the-fly computation one. In fact, only few papers about distortion correction or rectification on FPGA rely on this strategy \cite{a_passive_rgbd_sensor_for_accurate,real_time_rectification_for_stereo_correspondence,high_resolution_stereo_in_real_time}. The main goal of this paper is to highlight that, instead, the sampling approach is typically very accurate even with a small number of samples (\emph{i.e.}, large subsampling factor $n$) that require only a fraction of the whole LUT size, making it compatible with the amount of memory available even in cheaper FPGAs.

In the next section we thoroughly evaluate the performance, in terms of accuracy against increasing lens distortion, of each considered distortion correction approach providing a brief analysis of their resource utilization on FPGA.

\section{Evaluation of the Proposed Solutions}

In this section, the considered hardware solutions are analyzed in detail from two different perspectives:
\begin{enumerate}
	\item Geometric error: The root mean square error (RMSE) for each $(x, y)$ between the software generated map values and the FPGA solutions is computed and plotted considering different increasing lens distortions\footnote{Lens distortion factor $1$ corresponds to a minimal distortion, while distortion factor $5$ corresponds to the maximum distortion allowed by OpenCV 2.4 (almost equivalent to the distortion introduced by the short focal length of GoPro lenses). Unfortunately, given the limited number of pages, we are not able to detail this matter exhaustively.}
	\item Resources estimation: an estimate of the needed hardware resources for implementing the specific approach is given. Only the resources needed for the \emph{Map Module} depicted in Figure \ref{map-module} are considered (the other modules in Figure \ref{Design-Overview} are the same for all the considered approaches)
\end{enumerate}

\subsection{Complete On-the-fly Function Computation}

An important parameter to consider in this solution is the number of fixed point fractional digits that are exploited in the implementation of the \emph{on-the-fly} computation operators on hardware. Let us recall that a greater number of fractional digits leads on one hand to an improved accuracy of the final result, but on the other hand to a higher computational cost. This trade off is fundamental and therefore it required some further exploration. In particular, different values of fractional bits (12, 16, 20) have been considered and compared.

\begin{figure}[t]
	\centering
	\input{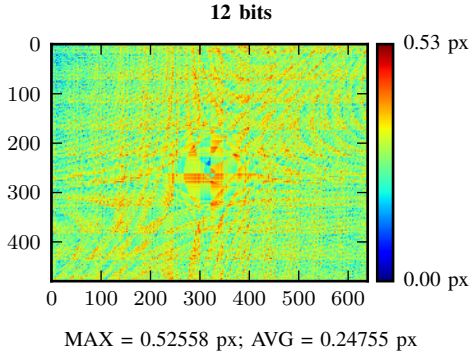}
	\caption{On-the-fly computation with 12 bits of fractional precision. Geometric RMSE, 480p resolution. Lens distortion factor = 3}
	\label{computation_480p_lens2_20bits}
\end{figure}

\begin{figure}[t]
	\centering
	\includegraphics[width=5.5cm]{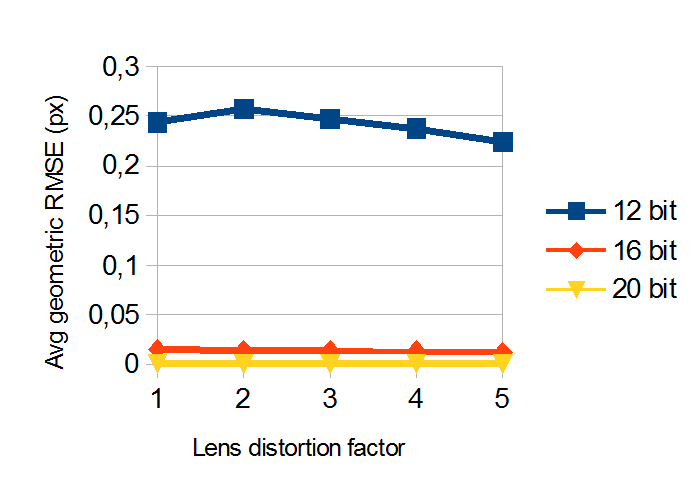}
	\caption{On-the-fly computation. Geometric error on the increasing distortion factor, 480p resolution. Different fractional precision values considered}
	\label{chart_rmse_computation_480p}
\end{figure}

From this analysis, two very positive aspects of this approach are noticed. The first one is that, with the exception of the 12 bits fractional precision, the average RMSE is quite low, which means that the FPGA map is almost equivalent to the floating-point software one also when considering strong lens distortion (Figure \ref{chart_rmse_computation_480p}). Other than that, the geometric error is almost uniformly distributed within the frame (Figure \ref{computation_480p_lens2_20bits}).

In order to understand the resource footprint of this solution, it is necessary to analyze the considered mapping function \cite{initundistortrectifymap} from an FPGA perspective. The total hardware cost estimate is shown in Figure \ref{number_of_operators} and discussed in the remainder.

\subsection{Map Sampling}

An important parameter to consider when evaluating this approach is the subsampling factor $n$, that is directly related to the total number of samples that are extracted from the original full resolution Look-up-Table: the bigger $n$, the smaller the number of samples. Sampling factors of 7 (128px), 6 (64px), and 5 (32px) have been considered during this study.

\begin{figure}[t]
	\centering
	\input{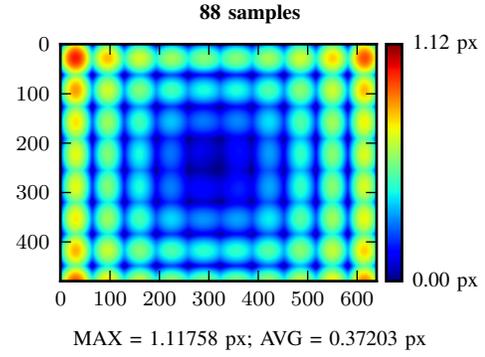}
	\caption{Map sampling with a sampling factor of 6. Geometric RMSE, 480p resolution. Lens distortion factor = 3}
	\label{sampling_480p_lens2_88samples}
\end{figure}

\begin{figure}[t]
	\centering
	\includegraphics[width=5.5cm]{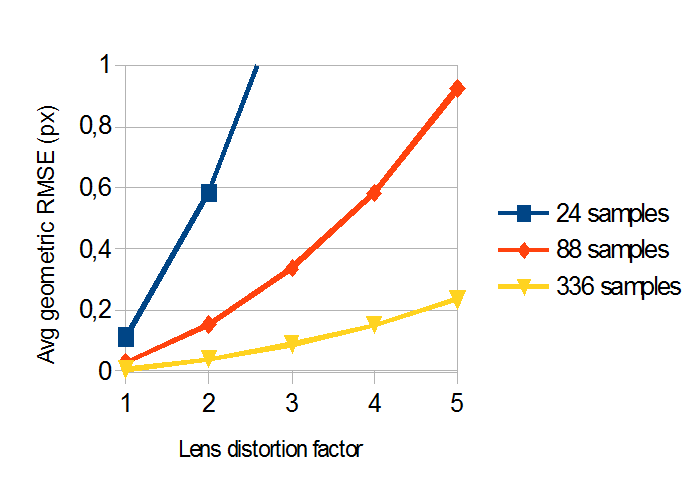}
	\caption{Map sampling. Geometric error on the increasing distortion factor, 480p resolution. Different sampling factor values considered. Fixed-point LUT samples are used (8 bit fractional part)}
	\label{chart_rmse_sampling_480p}
\end{figure}

Different aspects of this approach can be noticed. First of all, the error given by the bilinear interpolation of the samples causes a ``cushion'' effect (Figure \ref{sampling_480p_lens2_88samples}). Then, it can be noticed that the relationship between samples number and geometric error is inversely proportional: the more the used samples, the lower the error (Figure \ref{chart_rmse_sampling_480p}). Moreover, there is a strong dependency on the amount of distortion: an increment of the distortion leads to an increment of the error.

Nevertheless, a significant conclusion is that very good accuracy results are achieved even with a very small amount of samples (\emph{i.e.}, 336 for a VGA resolution image), also in the case of lenses characterized by severe distortion. When analyzing the RMSE reported in Figure \ref{chart_rmse_sampling_480p}, it is worth to consider that distortion correction maps estimated with camera calibration procedures are characterized by a accuracy that is in the range [0.1 - 0.5] px. Therefore the approximation introduced by this method is comparable with the precision of the map itself, hence this procedure does not introduce a fundamental precision limitation.

\begin{figure}[t]
	\centering
	\includegraphics[width=4.5cm]{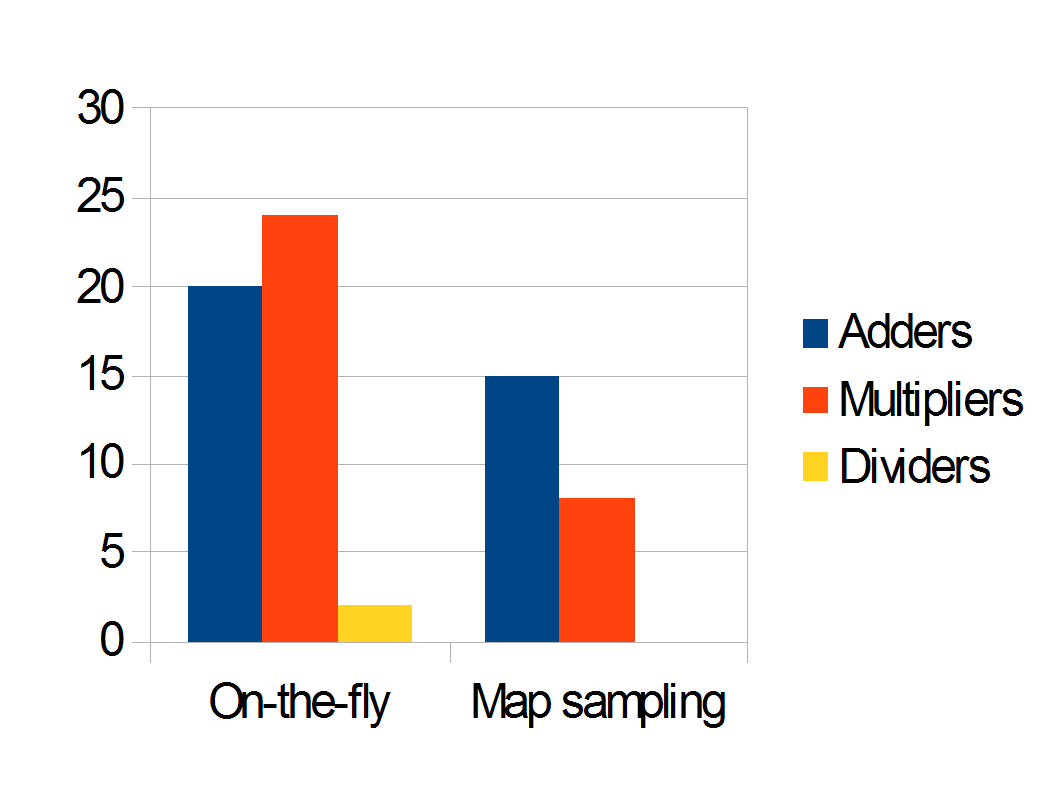}
	\caption{Number of hardware operators estimated for each approach. The number of operators needed in the map sub-sampling solution does not depend on the specific distortion model used or the chosen sampling factor}
	\label{number_of_operators}
\end{figure}

When considering the relative hardware costs, in this case it is not needed anymore to implement on hardware the map function, as a software pre-computed Look-up-Table of the function is used. Since the map is now sampled for fitting into static memory within the chip, an hardware implementation of a bilinear interpolator is needed for interpolating the samples and obtaining an approximation of the specific pixel's map value. It can be easily proven that a bilinear interpolation module can be implemented by using 3 multipliers, that means 6 total multipliers are needed for two interpolation modules (one for $map_x$ and another one for $map_y$). Along with some glue logic for address generation, an estimate of the needed resources is again shown in Figure \ref{number_of_operators}.

A very important achievement is that, with this solution, the amount of needed hardware resources for computation decreases significantly (if we also consider that dividers can be very expensive), and it is now independent on the mapping function complexity. Moreover, a small amount of memory, \emph{e.g.} $\sim 2.4 kb$ in case of VGA resolution with sampling factor 5 (32 px), is now needed for storing the map samples compared to the $\sim 2.46 mb$ of a full resolution LUT. Table \ref{table results} reports the actual resources usage of our specific implementation for the Xilinx Artix-7 architecture. The report is relative to a single image distortion correction module, using an 8px subsampling factor (very high accuracy) on VGA resolution (4941 total samples); the image buffer size is 50 image lines.

As theoretically analyzed before, this approach is definitely cheap in terms of hardware resources as well as in terms of memory footprint. The only significant hardware costs are related to the interpolation modules (three of them needed in total: two for the LUT values interpolation, one for the output pixel value computation).

Finally, observing Figure \ref{number_of_operators}, we can notice again that the subsampling method completely avoids expensive dividers, and yields a significantly lower amount of multipliers and adders, which is not depending anymore on the distortion model function. Moreover, by selecting appropriate parameters like the sampling factor, the subsampled approach leads to an accuracy that is suited for most practical applications.

% This command serves to balance the column lengths on the last page of the document manually. It shortens the textheight of the last page by a suitable amount. This command does not take effect until the next page so it should come on the page before the last. Make sure that you do not shorten the textheight too much.
%\setlength{\textheight}{20.2cm}

\section{Conclusions}

In this paper, we considered the image distortion correction problem for reconfigurable devices. Although most approaches in the literature rely either on the external memory solution or on the on-the-fly computation one, our analysis and evaluation shows that a good trade-off between the two approaches is represented by the map subsampling strategy. Moreover, compared to the on-the-fly solution, this approach decouples the hardware logic from the specific mapping function chosen, allowing flexibility in choosing different or more complex models with no need to change the architecture. A drawback of this approach is that, increasing the lens distortion factor, it might be not accurate enough in very particular setups including lenses with very short focal length (\emph{i.e.}, wide angle lenses).

\begin{table}[t]
\caption{Resources usage of our FPGA implementation of the overall image distortion correction module with map sub-sampling}
\small
\begin{center}
\begin{tabular}{|m{4cm}|m{1cm}|}
\hline
\textbf{Site Type} & \textbf{Used} \\ \hline
Slice LUTs & 475 \\ \hline
Slice Registers & 525 \\ \hline
Block RAM Tile & 16 \\ \hline
DSPs & 19 \\ \hline
\end{tabular}
\end{center}
\label{table results}
\vspace{-0.3cm}
\end{table}

% Although the on-the-fly approach is the most used among the current literature regarding image distortion correction on FPGA, often it is not the best solution. In fact, this approach is very expensive in terms of hardware resources, because it aims at providing such a high precision in the map values that is not needed in practice. The LUT subsampling approach should be instead preferred, since it is very cheap in terms of hardware resources and at the same time allows to get good precision, even when considering lenses characterized by high distortion.

%\section*{Acknowledgments}

\end{document}